\documentclass[journal]{IEEEtran}
\usepackage{amsmath,hyperref}
\usepackage{graphicx}
\usepackage{subfigure}
\usepackage{cite}
\usepackage{breqn}
\usepackage{amsfonts}
\usepackage{amsthm}
\usepackage{amsmath}
\usepackage{algpseudocode}
\usepackage{algorithm}

\usepackage[export]{adjustbox}
\usepackage{cleveref}
\usepackage{xcolor}
\usepackage{caption}
\usepackage{wrapfig}

\usepackage[disable]{todonotes}
\begin{document}
\title{Make Bipedal Robots Learn How to Imitate}

\author{Vishal~Kumar and~Sinnu~Susan~Thomas,~\IEEEmembership{Member,~IEEE}   
\thanks{Vishal Kumar and Sinnu S. Thomas are with School of Computer Science and Engineering, Digital University Kerala (IIITMK) Kerala 695317 India. e-mail: vishal.mi19, sinnu.thomas@iiitmk.ac.in}   
}

\markboth{}
{Kumar \textit{et al.}: Make Bipedal Robots Learn How to Imitate}

\maketitle
\begin{abstract}
Bipedal robots do not perform well as humans since they do not learn to walk like we do. In this paper we propose a method to train a bipedal robot to perform some basic movements with the help of imitation learning (IL) in which an instructor will perform the movement and the robot will try to mimic the instructor movement. To the best of our knowledge, this is the first time we train the robot to perform movements with a single video of the instructor and as the training is done based on joint angles the robot will keep its joint angles always in physical limits which in return help in faster training. The joints of the robot are identified by OpenPose architecture and then joint angle data is extracted with the help of angle between three points resulting in a noisy solution. We smooth the data using Savitzky-Golay filter and preserve the Simulatore data anatomy. An ingeniously written Deep  Q  Network (DQN) is trained with experience replay to make the robot learn to perform the movements as similar as the instructor. The implementation of the paper is made publicly available. 
\end{abstract}


\IEEEpeerreviewmaketitle

% ***********************
\section{Introduction}
\IEEEPARstart{H}{umans} learn to perform any specific task through various supervised, unsupervised, reinforcement, and semi-supervised techniques. Reinforcement learning is the way of learning by past experiences, for example consider a simulation environment in which a simulated agent will try to learn to play a game of chess, agent will have all the knowledge of the way chess pieces move and is called as action space for each iteration the agent will perform a move and the get reward as \textit{+1} if he cuts opponents piece, \textit{-1} if he land into dangerous situation and \textit{0} if nothing happens based on the reward and observation i.e. current position of all chess pieces agent will perform next move and slowly learns how to play chess and win. This kind of learning is more unsupervised, 
IL is a type of reinforcement learning where the learning agent tries to mimic the teacher. Human beings prefer IL \cite{Osa_2018} as we involuntarily employ it in our life irrespective of the success. We refer to the human or trainable agents (if we talk in reinforcement learning domain) gathering behaviour, expertise or way of completing any given task by observing an instructor or any other skilled agent doing it, this method of learning is getting popular because it eases teaching complex tasks without exhausting the resources. Number of possible scenarios in a complex application such as human-computer interaction \cite{booth1989introduction}, self-driving cars \cite{selfdriving}, augmented reality \cite{augmentedreality}, generative adversarial networks \cite{goodfellow2014generative}, and robot related tasks is vast and insane. Robots are classified according to their movements such as wheeled (uses wheels to move) \cite{KLANCAR20171}, tracked (uses tank like tracks to move) \cite{track2002}, quadruped (move on four legs) \cite{quadraped2014}, and biped (move on two legs) \cite{bipedal2009}. A lot of research is going into IL in bipedal robots \cite{SCHAAL1999233, Nehaniv2009ImitationAS} due to the reduced computational cost. If we consider an environment in which multiple agents are trying to climb to the top of the mountain, let us consider a random agent (Agent-22)with the ability of IL learn maneuvers that are likely to be useful for the task.
\iffalse as they are being used by some agents successfully in the same environment and the Agent-22 will be able to accomplish task with much ease.\fi Agent-22 learns by the reinforcement learning approach along with a referring database for agent to refer and adjust its actions to learn the task faster without an exploration and exploitation problem \cite{eande}. Bipedal robots utilize a human-like pair of legs to move around, they are different from humanoid robots \cite{humanoidrobot} as the bipedal may or may not have a complete human like torso, but humanoid must have. Cassie \cite{gong2018feedback} and Atlas \cite{Nelson2019} are examples of bipedal and humanoid robots respectively used for research. Making a bipedal learn any specific task is easy but the problem increases with a higher number of tasks. The domain increases so the variables on which the robot will depend would increase.
Training a bipedal to walk with human-like gait is a tedious task and we address this issue in this paper.
The contributions of the paper are as follows:
\begin{itemize}
    \item Extract joint angles for various movements of the bipedal robot using a single video.
    \item Train a DQN to predict the joint angles for the movement in OpenAI gym environment
\end{itemize}

We minimize the data reliability and large duration for training while extracting the joint angles data from a single video and then processing it to train DQN. The organization of the paper is as follows: Section \ref{Related_Works} discusses the state-of-the-art approaches in the field of IL, Section \ref{Proposed_Methodology} describes the working of the proposed approach, Section \ref{Experimental_Results} shows the experimental results and finally Section \ref{Conclusions} concludes the paper.

%%%%%%%%%%%%%RELATED WORK%%%%%%%%%%%%%%%
\section{Related Works}
\label{Related_Works}
Learning in humans is very diverse and subject to the task, we focus on implying IL to the bipedal robot POPPY \cite{DBLP:phd/hal/Lapeyre14}. One of the areas kept in mind while creating this bipedal is to provide researchers a low-cost machine to test their experiments. Poppy is a 25-degree of freedom (DOF) humanoid robot which is very less when compared to humans.
It has much lower DOF than humans who have 244 DOF but it mimics all the major joints a human will engage whenever performing a movement such as walking, sprinting, kicking. Whenever we humans move or perform a set of actions to complete a movement like sprinting we engage a lot of muscles, activate a lot of joints, apply forces and torque in varying manner to the body joints \cite{10.3389/fnsys.2016.00059} there are many variables that come into play, humans use active dynamics walking style i.e. while activating muscles for movement we consume energy in form of ATP (Adenosine triphosphate)\cite{atp} similarly in robots they consume power in form of electricity from battery, fuel cells etc. 

Bipedal robots utilise passive dynamic walking style \cite{passive} in which they do not consume any power while walking while simply going down the slope. This passive dynamic walking style saves energy but motion of bipedal is limited as it can only move in a specific direction and can not avoid obstacles. Ohta et al. \cite{fromactive} demonstrated a technique that can transform passive to active dynamic walking providing better control. Simulations for training a bipedal to walk is a tedious task, if we desire to make a bipedal learn to perform a movement with the help of neural networks like a simple feed-forward network (FNN) \cite{dai2017understanding} it is a grueling task for the network. IL \cite{Osa_2018, imitation} helps in this scenario as it is a supervised learning. In traditional supervised learning training data represents features and labels, whereas in IL training data demonstrate action and state. The state represents the position with respect to the world and velocities of the joints and action consist of a list of joint angles further the subject will perform in complete duration of movement in the scenario we are discussing. While training the agent mimics the movement performed by the subject or demonstrator. Learning in IL can be performed through mainly two classes behavioral cloning (BC) \cite{torabi2018behavioral,DBLP:conf/bmvc/GavenskiMGMB20} and inverse reinforcement learning (IRL) \cite{russellIL}. 

In BC, computer programs reproduce human behaviour or sub-cognitive skills of the human while performing a movement \cite{Sammut2010}, demonstrator perform the movement his or her action (joint angles and position) and the situation arise from that action (states) are recorded are used as input to train the agent, the agent will try to mimic the behaviour in learning process. BC is inefficient in complex tasks. In the IRL model of IL a reward function is being devised from the optimal behaviour of the demonstrator \cite{Sammut2010}, the devised reward function performance for training an agent will solely depend on how precise and clean movement is performed by the demonstrator. 

Humans use a lot of muscles while walking and mimicking but it is a difficult task in robots hence researchers develop a small bipedal robot \cite{1570404} and made it walk in forward direction with a constant speed using a single condition that robot gait must be efficient as human gait. Ames \cite{6709794} utilise human walking gait data to develop human-inspired control for walking of a bipedal robot while 57 muscles are engaged that lead to highly non-linear dynamics and forces acting on the muscles that sum up to hundreds of DOF. The authors created human-inspired controllers that utilizes time-based joint angles as output from human walking gait, processed and fed into a controller to control the robot. Learning algorithms were not used in this approach. Schuitema et al. \cite{1573573} use a simple reinforcement learning algorithm to develop a controller for a passive dynamic walking robot and to maintain the balance of the bipedal while walking distorted due to disturbances. A feedback control loop is used as a hip actuation mechanism and an imbalance of the bipedal calculates the variables included and provided to the controller. First the algorithm is tested while simulating and then the mechanical prototype is made and finally tested.

Dadashi et al. \cite{dadashi2021primal} proposed an algorithm where the primal Wasserstein distance between the expert and the agent state-action distributions is minimized, Ho and Ermon \cite{ho2016generative} use Shannon-Jensen divergence, and Fu et al. \cite{fu2018learning} use Kullback-Leibler divergence to match state-action distribution of agent to the expert. Using similarity measures for IL shows promising results but the practical implementation requires fine tuning of a lot of hyperparameters and parameters approximations. Wang et al. \cite{9009816} use Markov Decision Process \cite{sutton1998} for generation of pose prediction sequence of a human from an image. Ho and Ermon \cite{ho2016generative} present a generative adversarial network to train the agent for IL. The discriminator must be very fine tuned according to the problem in order to improve the performance of the algorithm. Christiano et al. \cite{christiano2017deep} propose a deep neural network based reinforcement learning from human preferences approach to perform the IL. This model can learn to imitate with minimal hours of human training data. Authors tested the performance of this method on Atari games and have satisfying results. Lepetit et al. \cite{pnp} use Perspective-n-points (PnP) to find camera-to-robot pose. This method works flawlessly for articulated robots but does not work for bipedal or high DOF robots and Lee et al. \cite{lee2020icra:dream} propose an algorithm to find joint positions of an articulated robot using a single image.

\iffalse
Thobbi and Sheng \cite{gestures} proposed a method to perform IL for hand gestures and and its evaluation for humanoid robots, algorithm aims to generalize over multiple demonstration of a single hand gesture and hence learn, it extracts features performs dynamic time wrapping lastly it apply weighted averaging to learn the hand gestures. Mechanical limitations of robot is not taken in consideration while performing this experiment so the gesture might get clipped if it extends beyond the limits of robot joints.
\fi
We propose to generate major joint angles with a single video of experts while performing movement with help of OpenPose architecture and then training a DQN to generate joint angles similar to the original input to make bot perform movement similar to expert demonstrator. While training we consider the physical limitations or joint limits of the robot so that the movement cannot be clipped. 

%%%%%%%%%%%%%METHODOLOGY%%%%%%%%%%%%%%%%%%%%%%%%%%%%%%%
\section{Proposed Methodology}
\label{Proposed_Methodology}
\begin{figure}[!htb]
    \centering
    \includegraphics[scale=0.40]{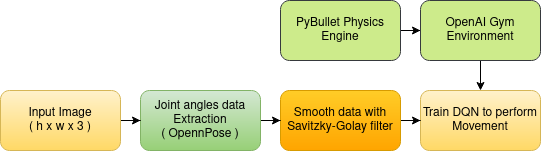}
    \caption{Flow chart of  the proposed approach.}
    \label{meth_odology}
\end{figure}

Fig. \ref{meth_odology} explains the approach where the joint angles data are extracted with the help of OpenPose \cite{cao2019}, smoothed after that with the help of DQN \cite{mnih2013playing}, and the robot is trained to perform the movement. 

\begin{figure}[!htb]
    \centering
    \includegraphics[width=0.5\textwidth]{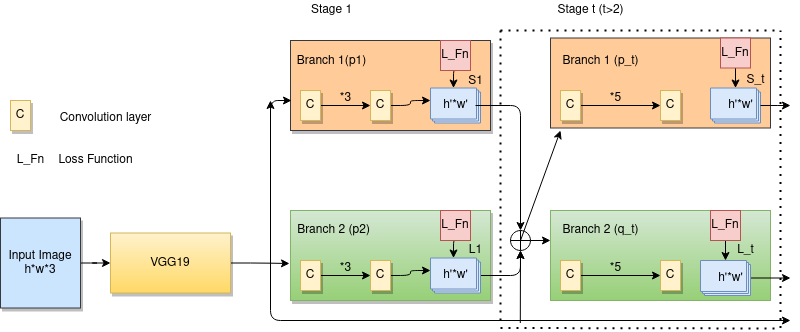}
    \caption{OpenPose Architecture}
    \label{open_pose}
\end{figure}

We use a color image as input to a network that produces 2D locations of the joints or anatomical keypoints of the human body such as [\textit{shoulder, elbow, knee, waist, head}]. These points are connected to form a $2D$ vector that represent limb with the help of PAFs (Part Affinity Fields) \cite{cao2017realtime} that encode the degree of union between points or joints. This is a bottom-up approach for detecting human pose in which, first the body parts are detected and then they are grouped together to form a complete human based on the PAFs. Fig. \ref{open_pose} shows the neural architecture for the OpenPose
.

Branch $1$ network produces PAF's $\textbf{P}^{1} = \theta^{1}(\textbf{IF})$ where $\theta^{1}$ refers to the neural network in branch 1 for summarizing in stage 1. In each successive stage original image features \textbf{IF} and predicted PAF's from previous are chained to produce precise predictions,
\begin{equation}
    \textbf{P}^{i}=\theta^{i}(\textbf{IF},\textbf{P}^{i-1}),  \forall\;1\leq i \leq \textit{L}^{p}
\end{equation} 
where $\textit{L}^{p}$ is total number of PAF stages in the network. After producing precisely refined PAF's prediction the process is repeated for confidence map detection starting from the most updated PAF prediction.
\begin{equation}
    \textbf{C}^{i}=\theta^{i}(\textbf{IF},\textbf{P}^{L^{P}}, \textbf{C}^{i-1}),  \forall\; 1\leq i \leq \textit{L}^{p}+\textit{L}^{c}
\end{equation}
where $\textit{L}^{c}$ is the total number of confidence map stages. $L_{2}$ loss is calculated between the estimated predictions and the ground truth maps and fields, loss functions for the PAF branch at stage $t_{p}$ and loss function of confidence map branch at stage $t_{cm}$ are:
\begin{equation}
    \textit{F}^{t_{p}}_{PAF} = \sum^{I}_{i = 1}\sum_{s}\textbf{B}(x,y).||\textbf{P}^{t_{p}}_{i}(x,y) - \textbf{P}^{*}_{i}(x,y)||^{2}_{2},
\end{equation}
\begin{equation}
    \textit{F}^{t_{cm}}_{CM} = \sum^{J}_{j = 1}\sum_{s}\textbf{B}(x,y).||\textbf{C}^{t_{cm}}_{j}(x,y) - \textbf{C}^{*}_{j}(x,y)||^{2}_{2}
\end{equation}
where $\textbf{P}^{*}_{i}$ is ground truth PAF and $\textbf{C}^{*}_{j}$ is ground truth confidence map and \textbf{B} is a binary mask with $\textbf{B}(x,y) = 0$ when the label is not present in the ground truth, the loss function is used to guide the network to predict the PAF's of body in branch 1 and confidence maps in branch 2 iteratively in precise manner. The overall loss function becomes:
\begin{equation}
    \textit{F}=\sum^{L_{p}}_{l=1}\textit{F}^{t_{p}}_{PAF}+\sum^{L_{p}+L_{c}}_{l=L^{p}+1}\textit{F}^{t_{cm}}_{CM}
\end{equation}
PAF preserves location and orientation of the detected keypoints and predict the connection between the limbs whether they are from same person or different person, each PAF is a 2D vector field for each limb in this 2D vector field a 2D vector encodes the direction of points from one point of limb to another. After extracting the keypoints or anatomically important joints of a person angle between the joints is calculated, output of the pose estimation network is locations of joints of human in sequence \textit{'nose', 'neck', 'r-shoulder', 'r-elbow', 'r-wrist', 'l-shoulder', 'l-elbow', 'l-wrist', 'r-hip', 'r-knee', 'r-ankle', 'l-hip', 'l-knee', 'l-ankle', 'r-eye', 'l-eye', 'r-ear', 'l-ear'} and locations of these keypoints is defined as,
\begin{equation}
    \textbf{K}_{pt}=(x_{i}, y_{i})
\end{equation}
where $x_{i}$ and $y_{i}$ are the pixel locations of the $\textbf{K}_{pt}$ keypoint from above mentioned joints. In this experiment we used \textit{'r-shoulder', 'r-elbow', 'r-wrist', 'l-shoulder', 'l-elbow', 'l-wrist', 'r-hip', 'r-knee', 'r-ankle', 'l-hip', 'l-knee', 'l-ankle'}. Shoulder joint angle is angle between a point vertically downwards to the shoulder joint and elbow joint for both left and right shoulder,  elbow joint angle is angle between shoulder and wrist for both right and left arm, hip joint angle is angle between a point vertically downwards hip joint and knee joint for both hip joints, knee joint angle is angle between hip joint and ankle joint for both left and right legs. We have three points $p_{1}=(x_{1}, y_{1}), p_{2}=(x_{2}, y_{2})$ and $p_{3}=(x_{3}, y_{3})$ and angle between three points is:
\begin{equation}
    \Theta = tan^{-1}\frac{m_{2}-m_{1}}{1+m_{1}m_{2}}
\end{equation}
where $m_{1} = (y_{2}-y_{1})/(x_{2}-x_{1})$ and $m_{2} = (y_{3}-y_{2})/(x_{3}-x_{2})$ are slopes of two line formed by joining $p_{1}$, $p_{2}$ and $p_{2}$, $p_{3}$ respectively. After all the major joint angles are recorded and saved, the joint values are used as input to an OpenAI gym \cite{Brockman2016} environment to make the bot learn and perform the movement. The environment has the Poppy Humanoid robot \cite{DBLP:phd/hal/Lapeyre14} as the agent, initially the shoulder joints of the robot are not in optimal standing human condition the arms were stretching away from the torso. Hence, we replaced all the RPY (roll, pitch and yaw) values to zero in the URDF ( Universal Robot Description Format ) file of the robot. Bot joint positions with respect to the world and joint forces are considered as state and the smoothed data of the joint angles obtained from above mentioned method are considered as actions.  Joint angles data is smoothed by applying Savitzky-Golay \cite{doi:10.1021/ac60214a047} filter using local least squares polynomial approximation without distorting signal tendency. Consider joint angles $[y_i], i=1, \ldots, n$ of the right shoulder over a period of $100$ milliseconds. The smoothed joint angle is given by the formula:-
\begin{equation}
    Y_{i} = \sum_{i=\frac{1-k}{2}}^{\frac{k-1}{2}} C_{z}y_{i+z}, \qquad \frac{k-1}{2}\leq i \leq n-\frac{k-1}{2}
\end{equation}
where $Y_{i}$ is smoothed data point, $k$ is the value of quadratic polynomial used for smoothing if $k=7$ then 7-point quadratic polynomial is used and $z = -3, -2, -1, 0, 1, 2, 3$ for $k=7$, $y_{i+z}$ is the data angle of index $i+z$ from original data and $C_{z}$ are convolutions coefficient and are the elements of the matrix
\begin{equation}
    C = (U^{T}U)^{-1}U^{T}
\end{equation}
where $U$ is a vandermonde matrix in which $i^{th}$ row of $U$ has values $1, z_{i}, z_{i}^{2}, . . . $. After smoothing the data we create a DQN that makes the bipedal robot POPPY learn to imitate the instructor. In DQN states are given as input to the neural network and Q-value of all possible action at that time is generated as output, as in reinforcement learning we do not know the target value there will be too much forking between the two if we use a single network to calculate predicted value and target value. To solve this issue we use two neural network one to approximate the Q-value functions and other as target network to estimate the target, however the architecture of target network is similar to the prediction network but with fixed weights the weights of the target network changes after every \textit{b} iterations (\textit{b}- a hyperparameter) to train smoothly as it will keep the target value fixed for a little while, weights are copied from the prediction network after \textit{b} iterations. Experience gained from the training is stored while training in the form of \textit{(state, action, reward, next-state)}.
\begin{table}[!htb]
	\begin{center}
		\begin{tabular}{|c|c|}
		\hline
		Number of Networks & 2 \\
		\hline
		Number of layers in each network & 5 \\
		\hline
		Activation Function & tanh/ReLU \\
		\hline
		Optimizer & Adam \\
		\hline
		Input Dimensions & 28 \\
		\hline
		Output Dimensions & 25 \\
		\hline
		Number of Hidden Layer Neurons & 768 \\
		\hline
		Learning Rate & 0.001 \\
		\hline
		Batch Size & 64 \\
		\hline
		Number of Epochs & 500 \\
		\hline
		\end{tabular}
		\caption{Neural Network Architecture and Hyperparameters}
		\label{hyperparams}
	\end{center}		
\end{table}
Algorithm \ref{alg1} shows DQN  \cite{AAAI1816976} to learn the task.
\begin{algorithm}
    \caption{DQN algorithm for IL}
    \label{alg1}
    \begin{algorithmic}[1]
        \State Initialize Replay Memory M of certain length\;
        \State Initialize prediction network with random weights $\Theta$\;
        \State Initialize target network with weights $\Theta^{*} = \Theta$\;
        \For{Z iterations}
        \For{\textit{p} iterations}
        \State Feed state to DQN and get Q-values for all possible actions
        \State Select action according to maximum Q-value
        \State Execute action and Store Experience as state, action, reward, next state
        \State Sample random batches from experience and calculate loss
        \State Loss function =$\mathcal{L}(Q)\;=\;\|Q_{Target}-Q_{Predicted}\|_2^{2}$
        \State Perform Gradient descent w.r.t network and minimize loss
        \EndFor
        \State Copy $\Theta \quad to \quad \Theta^{*}$
        \EndFor
    \end{algorithmic}
\end{algorithm}
We use two types of neural networks CNN and FNN for training and testing networks. The hyper parameters are given in Table \ref{hyperparams}, activation functions used are hyperbolic tan and ReLU by CNN and FNN respectively.

%%%%%%%%%%%%%%%%%%%%%RESULTS%%%%%%%%%%%%%%%%%%%%
\section{Experimental Results}
\label{Experimental_Results}
We develop an OpenAI Gym \cite{Brockman2016} environment with the help of PyBullet \cite{coumans2019} physics simulator to simulate the real world scenario while training the bot. Joint angles data is collected from a video of a person walking from \href{https://endlessreference.com/}{Endless Reference} with the help of OpenPose. The full implementation of the experiments in the paper is available \href{https://github.com/8-vishal/Imitation-Learning-in-bipedal-robots}{here}.

\begin{figure}[ht]
    \centering
    \includegraphics[scale=0.5]{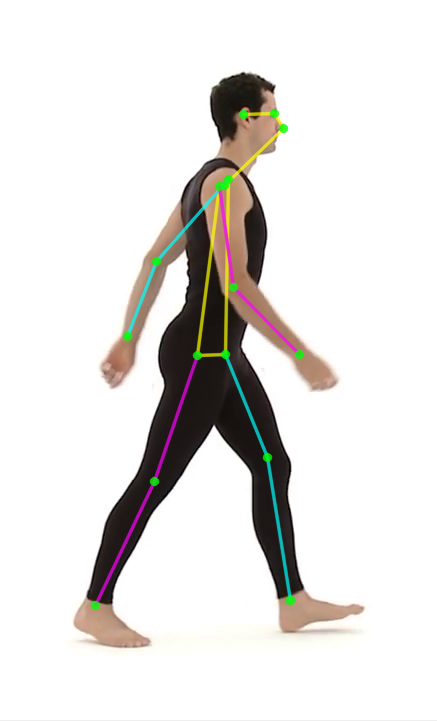}
    \caption{Detection of major joints and their natural limb connections with OpenPose Network.}
    \label{openpose_out}
\end{figure}
\begin{figure}[ht]
    \centering
    \includegraphics[scale=0.15]{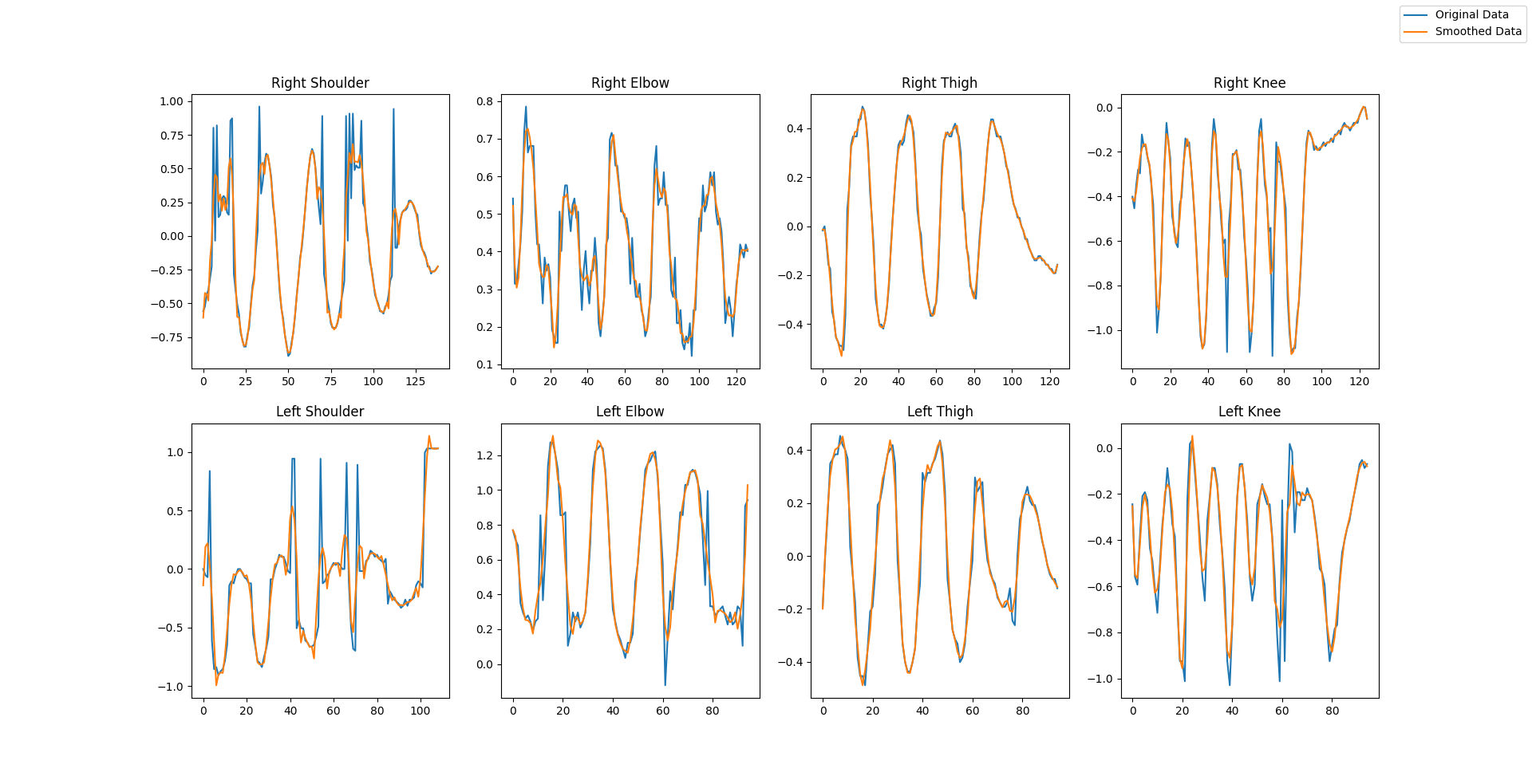}
    \caption{Original and smoothed data of adult male normal walk for all major joints.}
    \label{osdata}
\end{figure}
\begin{figure*}[ht]
    \centering
    \subfigure{\includegraphics[scale=0.25]{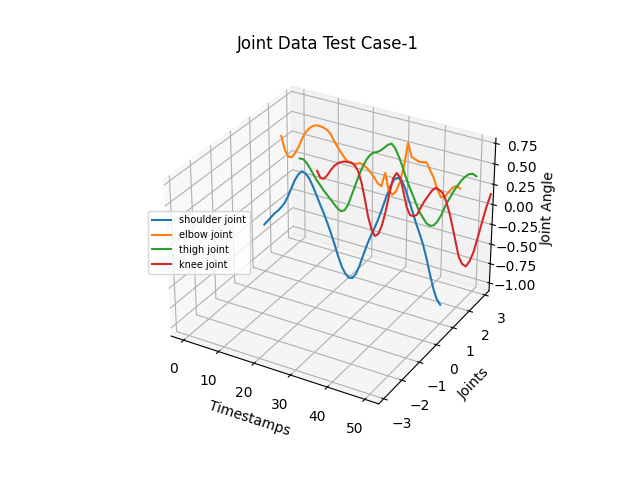}}
    \subfigure{\includegraphics[scale=0.25]{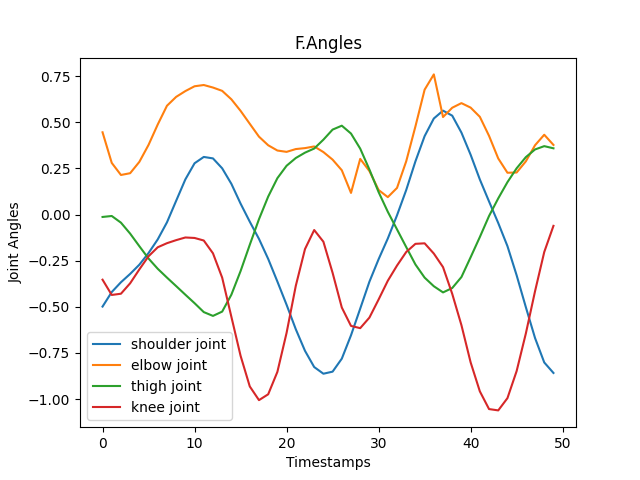}}
    \subfigure{\includegraphics[scale=0.25]{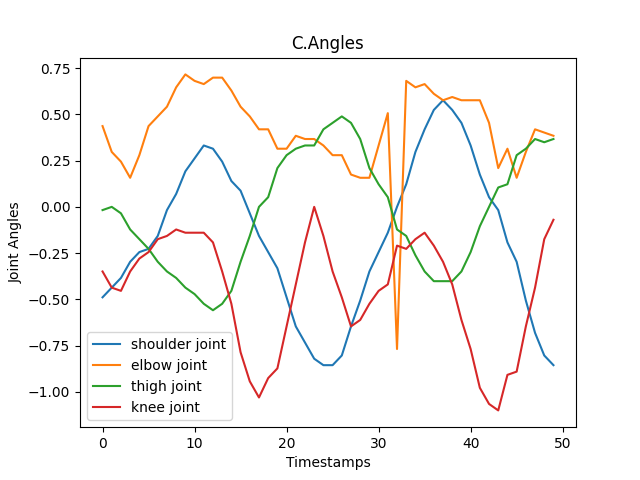}}
    \subfigure{\includegraphics[scale=0.25]{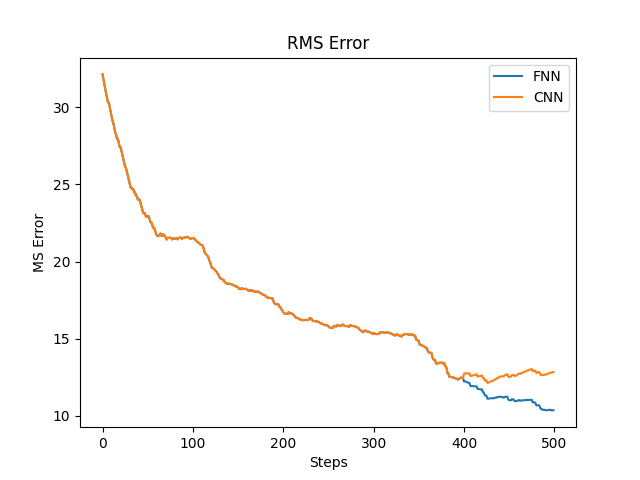}}
    \newline
    \subfigure{\includegraphics[scale=0.25]{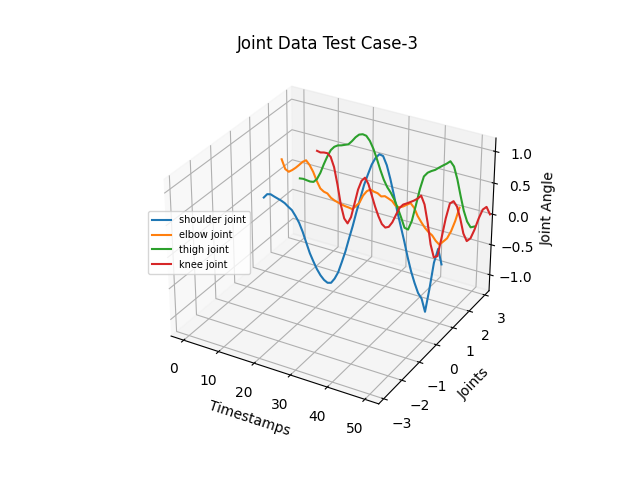}}
    \subfigure{\includegraphics[scale=0.25]{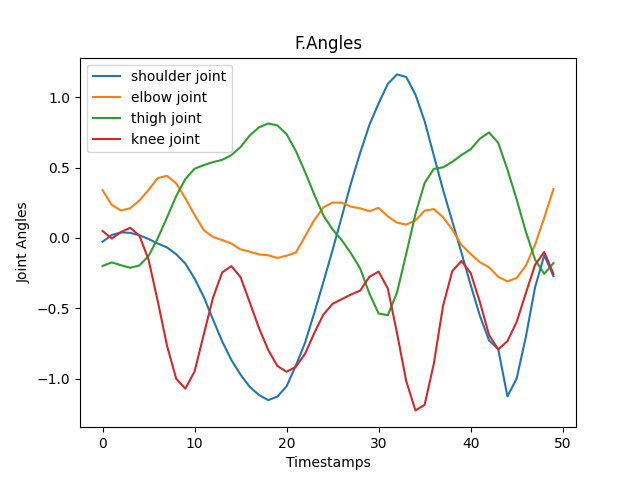}}
    \subfigure{\includegraphics[scale=0.25]{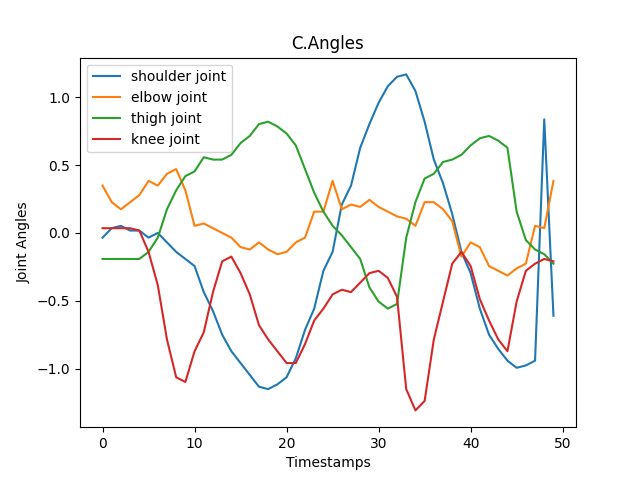}}
    \subfigure{\includegraphics[scale=0.25]{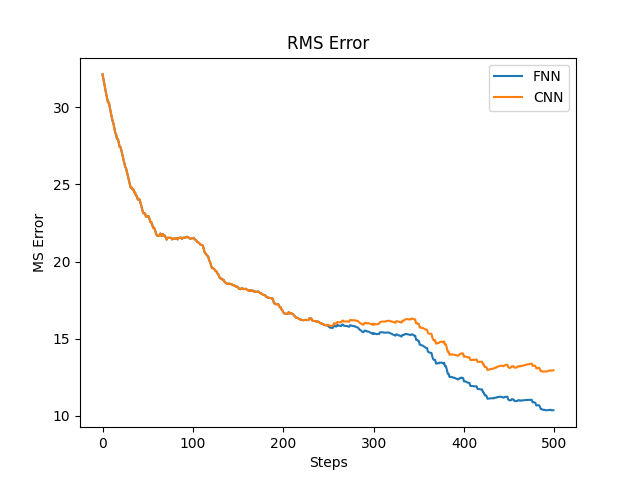}}
    \newline
    \subfigure{\includegraphics[scale=0.25]{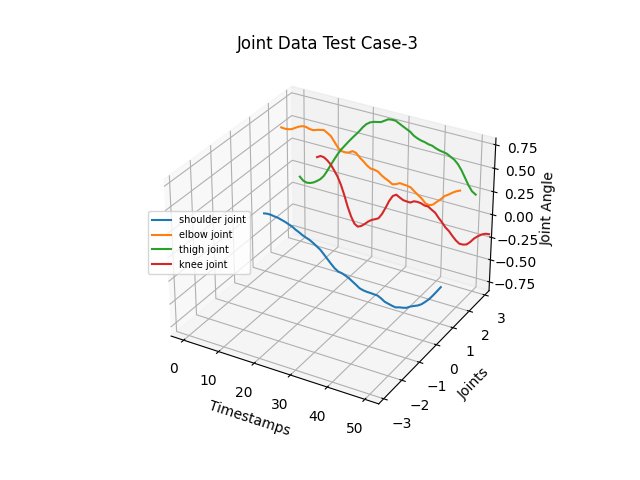}}
    \subfigure{\includegraphics[scale=0.25]{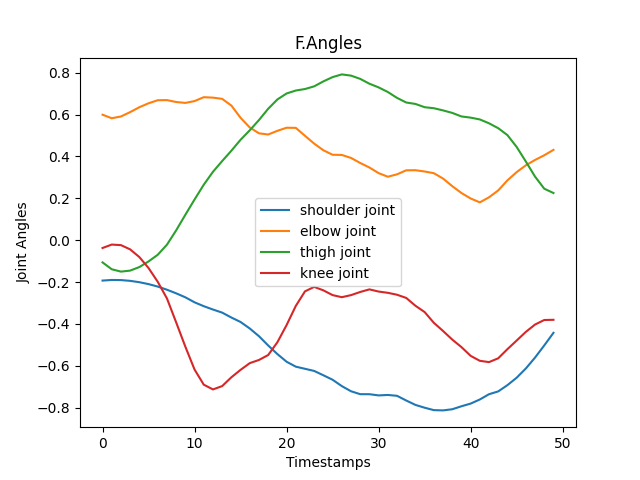}}
    \subfigure{\includegraphics[scale=0.25]{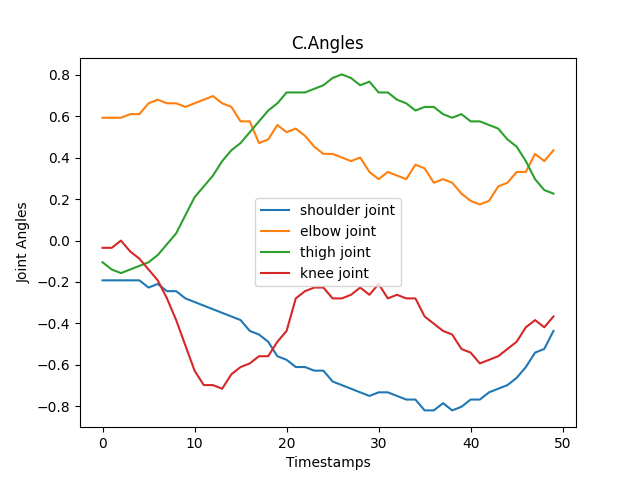}}
    \subfigure{\includegraphics[scale=0.25]{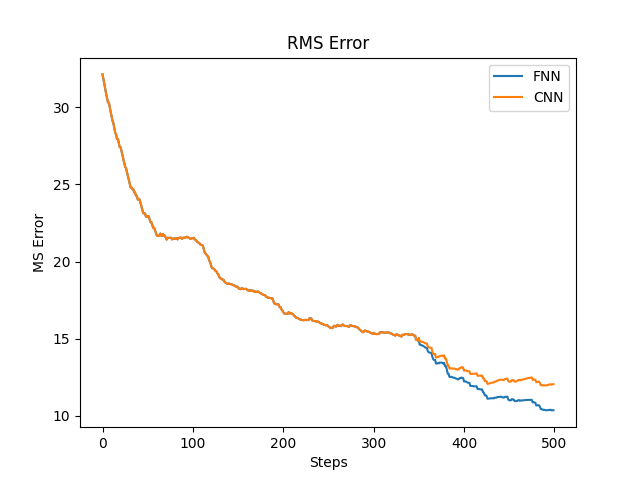}}
    \newline
    \subfigure{\includegraphics[scale=0.25]{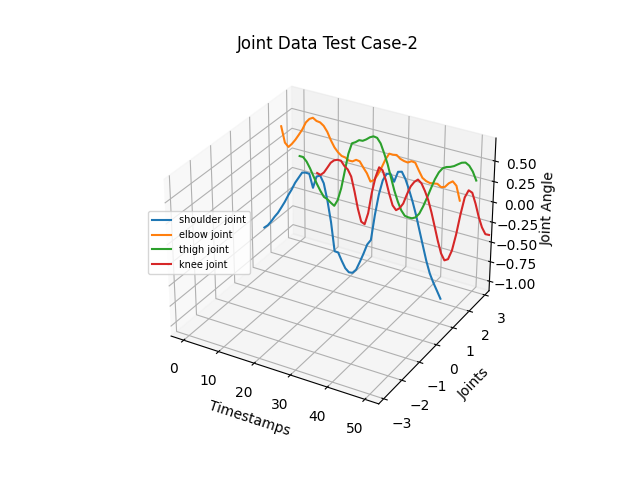}}
    \subfigure{\includegraphics[scale=0.25]{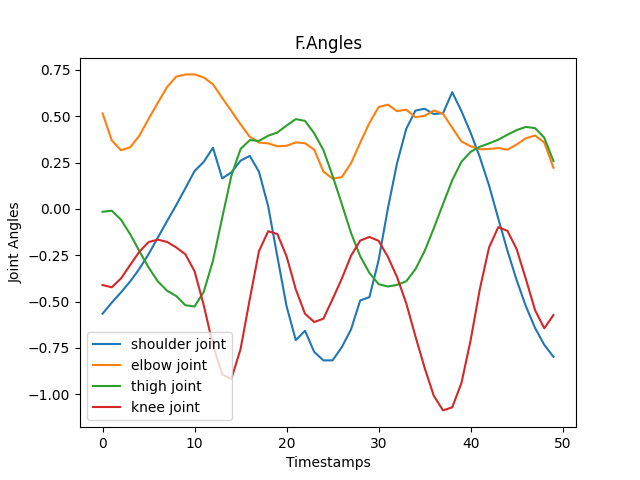}}
    \subfigure{\includegraphics[scale=0.25]{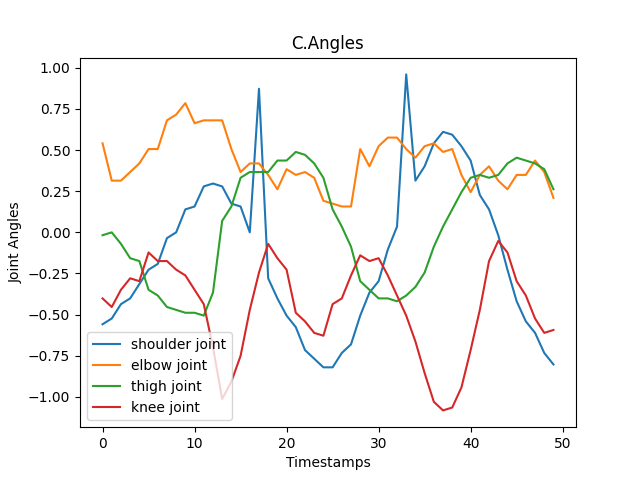}}
    \subfigure{\includegraphics[scale=0.25]{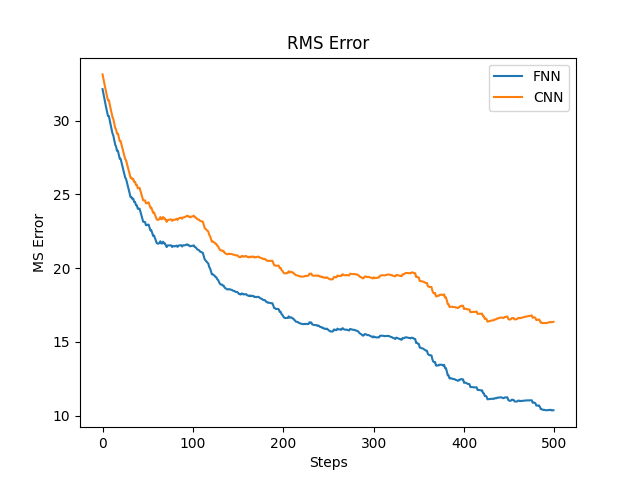}}
     \caption{First column shows 3-D plot of mean joint angles for all test cases used as input to train, second column shows output of FNN for mean joint angles for all test cases, third column gives output of CNN for mean joint angles for all test cases and fourth column shows the plot of RMS error for both the networks.}
    \label{mix}
\end{figure*}
Fig. \ref{openpose_out} shows the output of OpenPose with all major joints and connection between them as a skeleton and provides a noisy collected data because of some sudden movements of the instructor (person in video). We see absence of detected keypoints from the video so the data must be smoothed out as this data will lead to distortion in training of the agent. Savitzky-Golay filter is being used to smoothen the data, this filter will remove the sudden spikes in the data while keeping the data integrity as in Fig. \ref{osdata}.

Joint angles produced by these networks will be further known as F.Angles and C.Angles for output of FNN and CNN respectively.

\begin{table*}[ht]
	\begin{center}
		\begin{tabular}{|c|c|c|c|c|c|c|c|c|c|}
		\hline
		  & F.Shoulder & F.Elbow & F.Thigh & F.Knee & C.Shoulder & C.Elbow & C.Thigh & C.Knee \\
		\hline
		Case-1 & -0.15 & -0.18 & 0.01 & 0.41 & 1.23 & 1.03 & 2.15 & 1.96 \\
		\hline
		Case-2 & 0.17 & 0.21 & -0.14 & 0.36 & -1.36 & 2.00 & 1.98 & -0.75\\
		\hline
		Case-3 & 0.20 & -0.19 & 0.24 & 0.31 & 2.93 & 1.84 & 2.45 & 2.66\\
		\hline
		Case-4 & -0.13 & 0.16 & -0.19 & 0.53 & 2.88 & -1.53 & 2.26 & -2.98\\
		\hline
		\end{tabular}
		\caption{Average Mean angle error of joints}
		\label{meanerror}
	\end{center}
\end{table*}
\begin{table*}[ht]
	\begin{center}
		\begin{tabular}{|c|c|c|c|c|c|c|c|c|c|}
		\hline
		  & F.Shoulder & F.Elbow & F.Thigh & F.Knee & C.Shoulder & C.Elbow & C.Thigh & C.Knee \\
		\hline
		Case-1 & 1.80 & 1.54 & 1.33 & 1.40 & 4.57 & 2.54 & 6.86 & 4.02 \\
		\hline
		Case-2 & 2.08 & 1.92 & 3.47 & 2.12 & 12.90 & 12.81 & 9.51 & 11.28\\
		\hline
		Case-3 & 5.02 & 1.03 & 2.34 & 1.59 & 15.61 & 18.04 & 12.50 & 12.66\\
		\hline
		Case-4 & 6.03 & 1.96 & 2.17 & 2.53 & 22.88 & 21.03 & 21.26 & 22.98\\
		\hline
		\end{tabular}
		\caption{Euclidean distance between original and produced joint angle}
		\label{eucli_dean}
	\end{center}
\end{table*}
Fig. \ref{mix} shows four test cases to validate the proposed approach. We looked at (1) Adult male walking in normal condition, (2) Adult female walking in normal condition, (3) Adult male walking with long strides, and (4) Adult female walking with long strides. First column of Fig. \ref{mix} shows the original input joint angles used for training for all four test cases, second column shows plot of F.Angles, third column shows plot of C.Angles and the last column shows plots of root mean square (RMS). We conclude from Fig. \ref{mix} that the FNN predicts joint angles more accurately than CNN. The plots in the second column are smoother when compared to the plots in the third column and similar to the original input. When we compare a small timestamp from both second and third column plots we see that joint angle transitions are smooth and there are no fluctuations whereas in third column plots joint angle transitions have fluctuations which lead to shivers in output movement.

Average mean angle error is the difference of average of the predicted left and right joint angles of the body and average of training input of left and right joint angles. Positive average mean angle error depicts the predicted joint angles are smaller or equal to the original joint angles and vice versa. Table \ref{meanerror} shows average mean angle error of joints. While analysing the data in Table \ref{meanerror} it seems that the values of \textit{F.Shoulder, F.Elbow, F.Thigh, F.Knee} are much closer to zero which means the joint angle prediction is approximately equal, while values for \textit{C.Shoulder, C.Elbow, C.Thigh, C.Knee} are far from zero compared to F. values. 

Table \ref{eucli_dean} shows the Euclidean distance between the predicted joint angles and the original joint angles that is used for training for both the networks. F. values have lower euclidean distance which means the input and predicted are very much similar but for the C. values the euclidean distance has higher values which shows that the input and predicted are distant from each other.
\begin{figure}[!htb]
    \centering
    \includegraphics[scale=0.15]{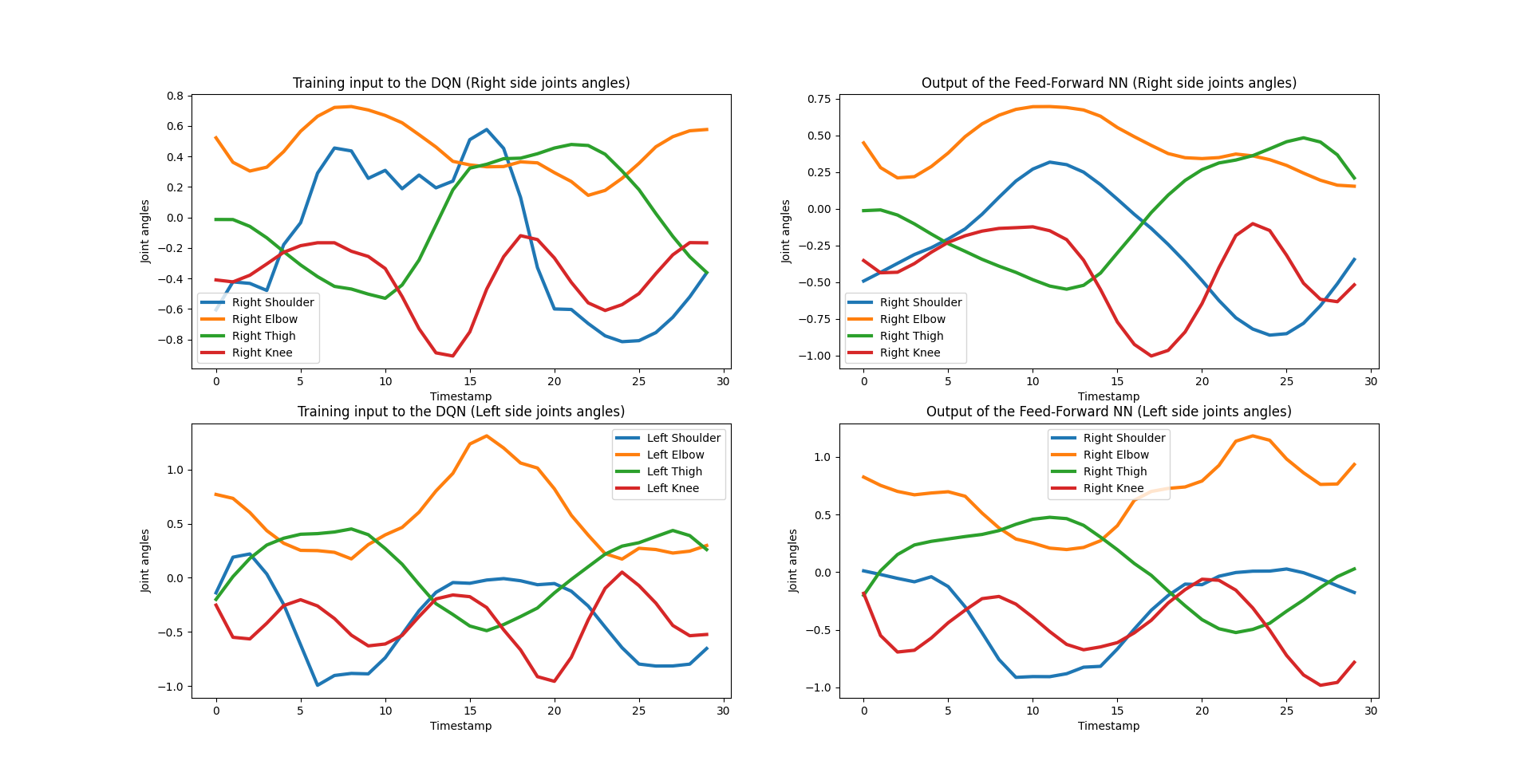}
    \caption{Right and Left side joint angles output from FNN network}
    \label{fnn_output}
\end{figure}
\begin{figure}[!htb]
    \centering
    \includegraphics[scale=0.15]{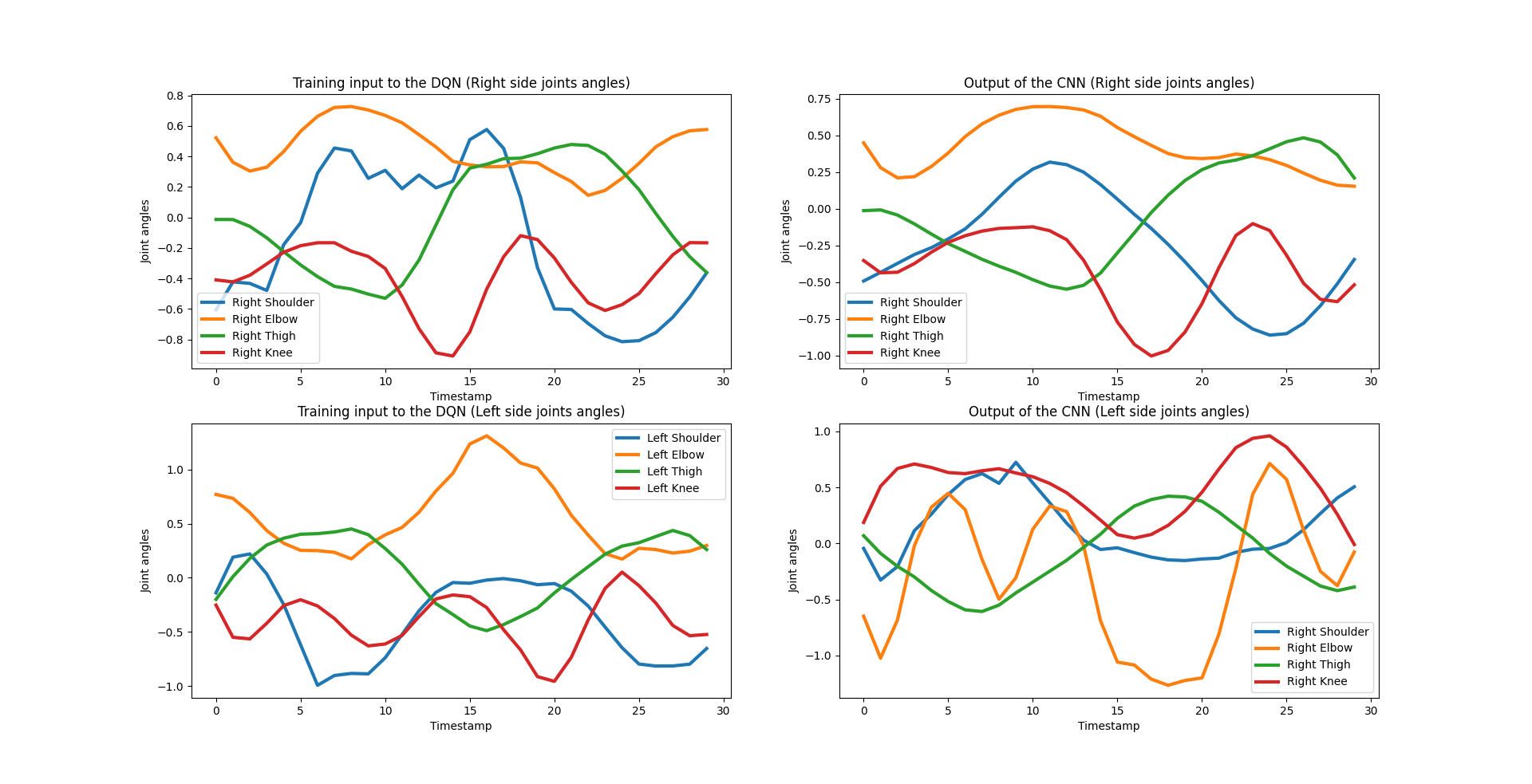}
    \caption{Right and Left side joint angles output from CNN network}
    \label{cnn_output}
\end{figure}

\begin{figure}[!htb]
    \centering
    \includegraphics[scale=0.25]{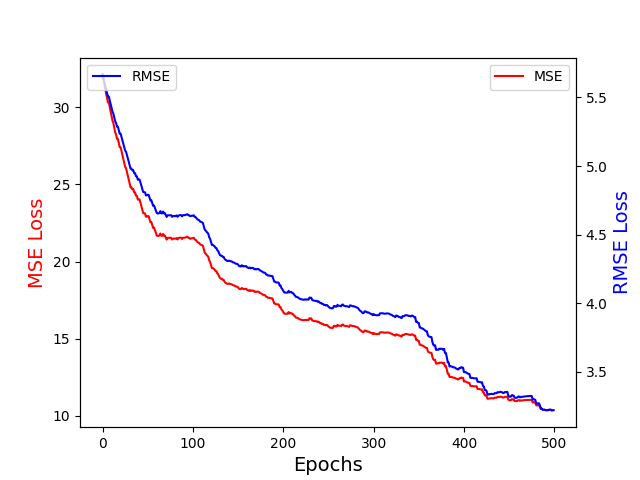}
    \caption{Loss with respect to epochs for proposed approach.}
    \label{msermse}
\end{figure}

We consider joint angles as the sole variable to perform IL due to unavailability of the desired data. Fig. \ref{fnn_output}  and Fig. \ref{cnn_outpu} shows the plot of joint angles produced by FNN  and CNN respectively after training for both the left and right side of the body. The left part of the plot is the input joint angles and right part of the graph is output joint angles. While comparing these two figures it seems that the output of F.Angles are much similar to the input angles used to train compared to the C.Angles as shown in the bottom left plot of Fig. \ref{cnn_outpu}. Fig. \ref{mix}, \ref{fnn_output}, and \ref{cnn_outpu} shows that the joint angles are totally different from the input angles. The prediction of F.Angles is comparatively similar to the original input data than C.Angles. This also shows that FNN performs better than CNN in this scenario. There will be some scenarios when we have to feed the input as an image instead of joint angles data in those cases CNN may perform better than FNN. Fig. \ref{msermse} shows the mean squared error (MSE) and root mean squared error (RMSE) for the training of both the networks, MSE loss is plotted on left side of $y$ axis and RMSE loss is plotted on right side of $y$ axis due to range variation of both losses.
The curves in Fig. \ref{msermse} shows that the loss is decreasing over a number of epochs and there is no sudden drop in loss which means the algorithm is learning instead of remembering the demonstrated action.

%%%%%%%%%%%%%%%%%%%%CONCLUSION%%%%%%%%%%%%%%%%%%%%%%%%%%%%%%%%%%
% ***********************
\section{Conclusions}
\label{Conclusions}
In this paper we made an agent learn to perform specific movement as demonstrated by the instructor using IL, first data generation is done by OpenPose, a video is fed to the OpenPose network frame by frame and the joint angles are extracted and stored in text files after this step data smoothing is done by Savitzky-Golay filter. Data is then fed to a DQN with a target network made to predict the values of original joint angles data. We used two networks for training and inferred that the performance of FNN performs better to give joint angles similar to the original data. The proposed approach works well for the slow movements such as walking and jogging in a computationally cost effective manner using a single video as the training data. In future with 3D-Data generation and other network architecture more complex movements similar to the instructor can be imitated. The data generated would be 2D joint angles and their position in 2D space. 3D data generation would constitute joint angles, positions, and velocity of joints in 3D space giving rise to more realistic training. 

%%%%%%%%%%%%%%%%%%%%%%%%%%%%%%%REFRENCES%%%%%%%%%%%%%%%%%%%%%%%%%%%%%%%%
\bibliographystyle{IEEEtran}
\bibliography{references}

\end{document}